\pdfoutput=1

\documentclass[11pt]{article}

\usepackage[final]{acl}
\usepackage{times}
\usepackage{latexsym}

\usepackage[T1]{fontenc}

\usepackage[utf8]{inputenc}

\usepackage{microtype}

\usepackage{inconsolata}

\usepackage{hyperref}
\usepackage{url}

\usepackage[most]{tcolorbox} 
\usepackage{graphicx}
\usepackage{subfigure}
\usepackage{wrapfig}

\usepackage{amsmath}
\usepackage{amstext}
\usepackage{amsfonts}
\usepackage{bm}

\usepackage{bbm}

\usepackage{alltt}
\usepackage{xcolor}
\usepackage[table]{xcolor}
\newtcolorbox{AIBox}[2][]{aibox,title=#2,#1}

\tcbset{
  aibox/.style={
    width=1.0\textwidth,
    top=5pt,
    colback=black!05,
    colframe=black!20,
    colbacktitle=black!50,
    enhanced,
    center,
    attach boxed title to top left={yshift=-0.1in,xshift=0.1in},
    boxed title style={boxrule=0pt,colframe=white,},
  }
}

\usepackage{multirow}
\usepackage{booktabs}
\usepackage{array}
\usepackage{caption}
\usepackage{multirow}
\usepackage{booktabs}
\usepackage{array}
\usepackage{caption}
\usepackage{color}
\usepackage{colortbl}
\usepackage{tablefootnote}
\usepackage{adjustbox}

\usepackage{caption}
\usepackage{algorithm}
\usepackage{algpseudocode}

\algnewcommand{\LineComment}[1]{\Statex ~~~~~~\textsc{//}~\textit{#1}}

\usepackage{enumitem}
\setenumerate[1]{itemsep=0pt,partopsep=0pt,parsep=\parskip,topsep=5pt}
\setitemize[1]{itemsep=0pt,partopsep=0pt,parsep=\parskip,topsep=5pt}
\setdescription{itemsep=0pt,partopsep=0pt,parsep=\parskip,topsep=5pt}

\usepackage{soul}

\usepackage{tikz}
\usepackage[edges]{forest}
\definecolor{hidden-draw}{RGB}{64,101,149}
\definecolor{hidden-pink}{RGB}{231,239,250}

\usepackage[mathscr]{euscript}
\newcommand{\dataset}{\textsc{StripCipher}\xspace}

\usepackage{amssymb}  
\usepackage{pifont}

\usepackage{scalerel}
\usepackage{graphicx}  
\usepackage{array}     
\usepackage{booktabs}  
\usepackage{ragged2e}  

\usepackage{xspace}

\title{
\texorpdfstring{\includegraphics[width=18pt]{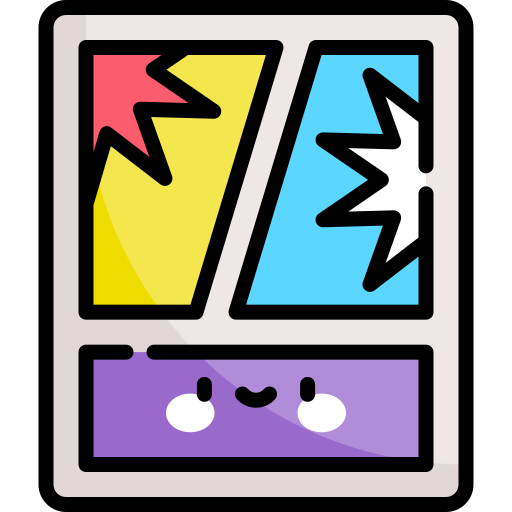}}{}
\textit{Beyond Single Frames:} Can LMMs Comprehend  Implicit Narratives in Comic Strip?
}

\author{%
\textbf{Xiaochen Wang}\textsuperscript{1 $*$},
\textbf{Heming Xia}\textsuperscript{2 $*$},
\textbf{Jialin Song}\textsuperscript{1 $\dagger$},
\textbf{Longyu Guan}\textsuperscript{1 $\dagger$}, \\
\textbf{Qingxiu Dong}\textsuperscript{1}, 
\textbf{Rui Li}\textsuperscript{1},   
\textbf{Yixin Yang}\textsuperscript{1},
\textbf{Weiyao Luo}\textsuperscript{1}, 
\textbf{Yiru Wang}\textsuperscript{4}, \\
\textbf{Yifan Pu}\textsuperscript{3}, 
\textbf{Xiangdi Meng}\textsuperscript{1}, 
\textbf{Wenjie Li}\textsuperscript{2}, 
\textbf{Zhifang Sui}\textsuperscript{1 $\ddagger$} \\
\textsuperscript{1} State Key Laboratory of Multimedia Information Processing, Peking University\\
\textsuperscript{2} Department of Computing, The Hong Kong Polytechnic University\\
\textsuperscript{3} Tsinghua University \quad
\textsuperscript{4} ModelTC \\
  \texttt{ wangxiaochen@stu.pku.edu.cn}
}

\begin{document}
\maketitle

\renewcommand{\thefootnote}{\fnsymbol{footnote}}

\footnotetext{\textsuperscript{$*$} Equal first contribution. } 
\footnotetext{\textsuperscript{$\dagger$} Equal second contribution.}
\footnotetext{\textsuperscript{$\ddagger$} Corresponding author.}

\begin{abstract}
Large Multimodal Models (LMMs) have demonstrated strong performance on vision-language benchmarks, yet current evaluations predominantly focus on single-image reasoning. 
In contrast, real-world scenarios always involve understanding sequences of images. A typical scenario is comic strips understanding, which requires models to perform nuanced visual reasoning beyond surface-level recognition.
To address this gap, we introduce \dataset, a benchmark designed to evaluate the model ability on understanding implicit narratives in silent comics. \dataset is a high-quality, human-annotated dataset featuring fine-grained annotations and comprehensive coverage of varying difficulty levels.
It comprises three tasks: visual narrative comprehension, contextual frame prediction, and temporal narrative reordering.
Notably, evaluation results on \dataset reveals a significant gap between current LMMs and human performance---e.g., GPT-4o achieves only 23.93\% accuracy in the reordering task, 56.07\% below human levels.
These findings underscore the limitations of current LMMs in implicit visual narrative understanding and highlight opportunities for advancing sequential multimodal reasoning.

\end{abstract}

\section{Introduction}

\begin{quote}
    \textit{In the space between the panels, human imagination takes separate images and transforms them into a single idea.}\par\raggedleft--- Scott McCloud (1993)
\end{quote}

\begin{figure}[t]
\centering
\includegraphics[width=0.95\columnwidth]{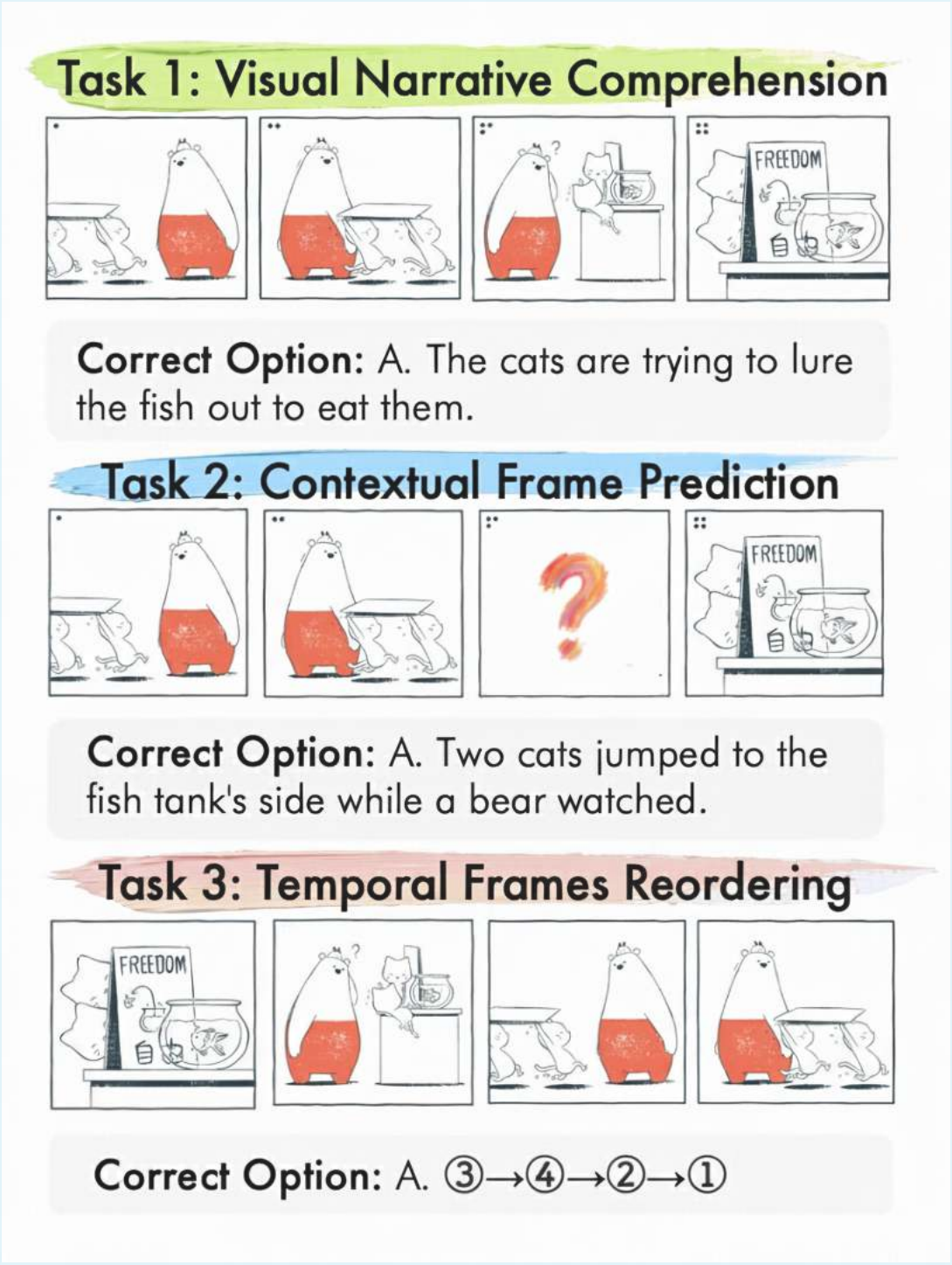}
\caption{An example of three tasks:  prediction, comprehension, and reordering from the \dataset dataset. All tasks are presented as multiple-choice questions, with distractors excluded due to limited context.}
\label{fig:intro}
\vspace{-5mm}
\end{figure}

Comics are a widely accessible medium, appealing to audiences ranging from children to adults.
Sophisticated comics heavily harness context cues, cultural references, and visual metaphors to convey layered, implicit meaning~\cite{magnussen2000framework,manning1998understanding,pressman2014digital}. Generally, children with basic comprehension skills can grasp the logical relationships and underlying meanings of multi-panel comics.


LMMs have demonstrated impressive performance across various visual-language tasks such as image captioning~\cite{liu2023llava, Ghandi:2024imagecaption}, text recognition~\cite{ocrbench}, visual question answering~\cite{Lu:2023vqa, Zhu:2024minigpt-4} and video understanding~\cite{Zhang:2023video-llama, maaz:2024video-chatgpt}. 
 However, the question remains whether these models can truly grasp the implicit narratives embedded in comics—a challenge that goes beyond conventional OCR and text reasoning. 
 Existing comic benchmarks~\cite{huang2016visual,hong2024summary,surikuchi2024not,vivoli2024comix,sachdeva2024manga} have predominantly focus on instance level tasks like object detection and dialogue generation in text-heavy comics, allowing LMMs to rely on OCR and superficial text processing rather than deep visual comprehension. Recent benchmarks~\cite{yang2024:deepsemantic, liu2024:iibench} evaluate single-frame comics with an emphasis on surface-level interpretation, neglecting the sequential dependencies of multi-frame narratives.

 Motivated by these limitation, we focus on silent comic strips—compositions devoid of explicit textual dialogue and consisting of 3–8 frames. The transition from single images to sequential data is not just a logical progression—it mirrors real-world perceptual processes and aligns with emerging trends in multimodal research. Video-based LLMs~\cite{zhang2025videollama,maaz:2024video-chatgpt,zhang2024llavanextvideo} also preprocess video input by sampling key frames before performing inference. 
 
 Multi-frame comics occupy a distinctive niche between static images and dynamic videos, conveying complete narratives through a minimal, densely packed sequence of frames. Much like keyframes in video analysis, these structured sequences create a controlled yet challenging environment for evaluating temporal understanding, causal relationships, and contextual reasoning in LMMs.

We introduce \dataset, a novel benchmark designed to assess the implicit temporal-visual reasoning ability of LMMs. \dataset consists of three challenging tasks: (1) contextual frame prediction task to eval visual contextual reasoning, (2) visual narrative comprehension task to infer overall implicit narrative, moving beyond local cues (3) temporal frames reordering, as shown in Figure~\ref{fig:intro}. 
We utilized GPT-4o to preprocessing and generate answers and distractors, then employed human annotators double-verify and remove low-quality entries, resulting in 2,170 samples. The dataset includes classic comics like "Peanuts" as well as recent published "Buni", ensuring diversity.

Our comprehensive evaluation of $16$ state-of-the-art LMMs on \dataset reveals a substantial performance gap compared to human capabilities in sequential image comprehension, especially in the frame reordering task. Most notably, GPT-4o achieves only $23.93\%$ accuracy in the reordering subtask and trails human performance by $30\%$ in visual narrative comprehension. Further quantitative analysis identifies several key factors affecting the sequential understanding performance of LMMs, highlighting the fundamental challenges for future LMM development.

\begin{table}[t]
\centering
\small
\setlength{\tabcolsep}{3pt}
\renewcommand{\arraystretch}{1.2}
\resizebox{0.48\textwidth}{!}{

\begin{tabular}{l l c  r r r}
\toprule
\textbf{Benchmark} & \textbf{Task} & \textbf{Seq.} &  \textbf{\#Fra} & \textbf{\#Seq} & \textbf{\#Cat} \\
\midrule
HCD & Funniness Classification & \ding{55}  & 50 & 0 & 1 \\
MTSD & Sarcasm Classification & \ding{55}  & 9,638 & 0 & 3 \\
HUB & Matching, Ranking, Explanation  & \ding{55} & 651 & 0 & 1 \\
MORE & Sarcasm Explanation & \ding{55}  & 3,510 & 0 & 1 \\
DEEPEVAL & Description, Title, Deep Semantics & \ding{55}  & 1,001 & 0 & 6 \\
AutoEval-Video & Open-Ended Question Answering & \ding{51}  & 1,033 & 327 & 12 \\
Mementos & Description Generation & \ding{51} &  8,124 & 699 & 1 \\
\midrule
\dataset{} & Prediction, Comprehension, Reorder & \ding{51}  & 3,660 & 896 & 6 \\
\bottomrule
\end{tabular}
}
\caption{Features and statistics of \dataset{} and related datasets. \texttt{Seq.} indicates whether image sequences are included. \#Fra denotes the number of frames (sub-images), \#Seq is the number of image sequences, and \#Cat is the number of image categories.}
\label{tab:Features}
\vspace{-5mm}
\end{table}



Our key contributions are as follows:
We introduce \dataset, the first benchmark that evaluates LMMs on implicit visual narrative reasoning with silent comics.
We propose three novel tasks (narrative comprehension, frame prediction, and reordering) that challenge existing LMMs beyond single-frame comic understanding.
We comprehensive evaluat 16 state-of-the-art LMMs, revealing substantial gaps between AI and human capabilities.
\begin{figure*}[t]
\centering
\includegraphics[width=0.95\textwidth]{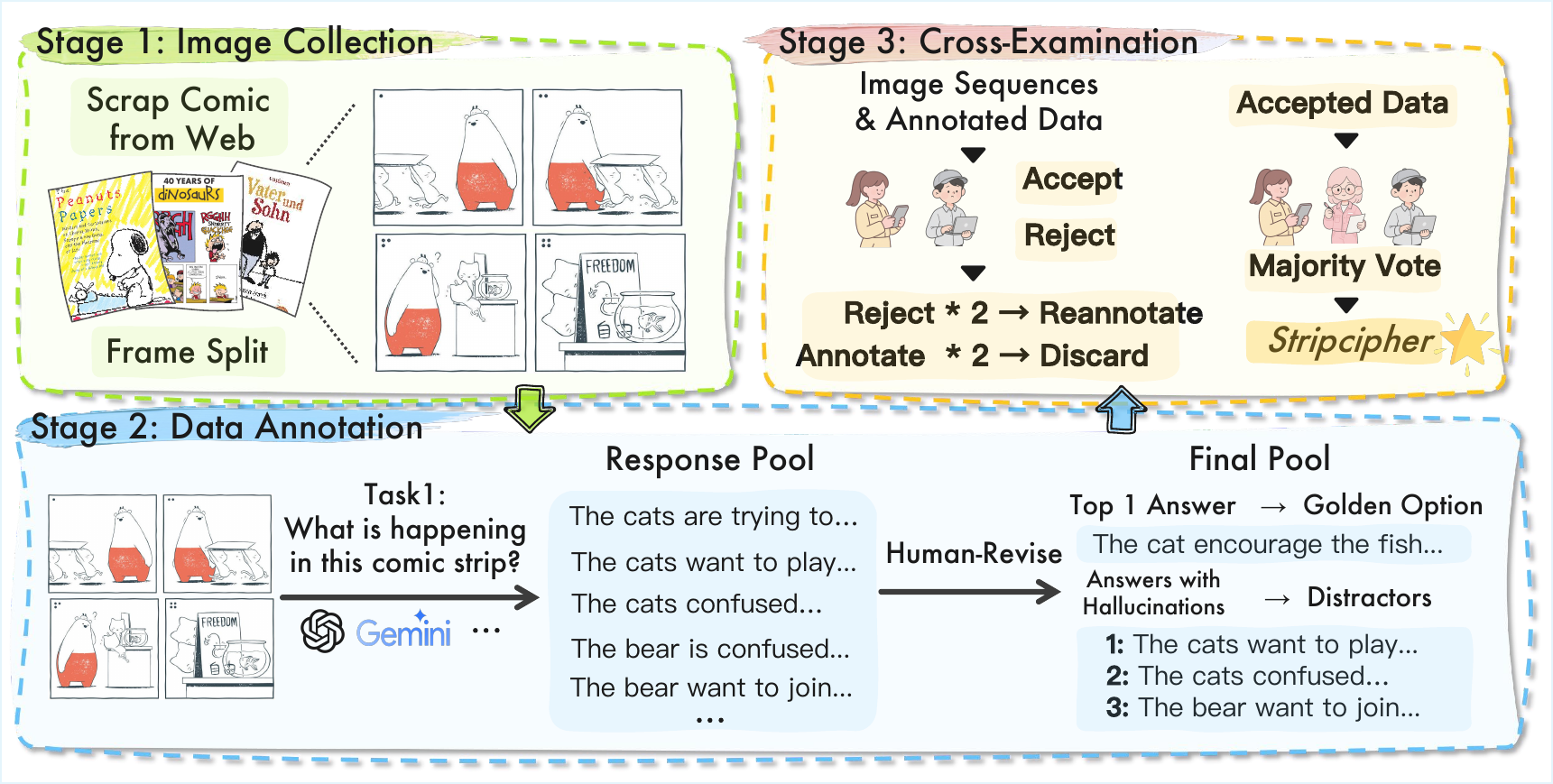}
\caption{Schematic diagram of \dataset{} dataset construction process including three stages: \textit{Image Collection}, \textit{Data Annotation} and \textit{Cross Check}. Only comprehension task is displayed, as Prediction follows the same process.}
\label{fig:construct}
\vspace{-5mm}
\end{figure*}

\section{Related Work}

\paragraph{Large Multimodal Models}
LLMs have demonstrated exceptional performance in various natural language understanding and generation tasks~\cite{dubey2024llama, liu2024deepseek, ray2023chatgpt}. Building on the scaling laws of LLMs, the next generation of LMMs has emerged, utilizing LLMs as their backbone. Several closed-source LMMs~\cite{reid2024gemini, driess2023palm, yang2023dawn}, including GPT-4o~\cite{hurst2024gpt40}, have shown remarkable capabilities in handling complex multimodal inputs~\cite{ fu2023mme, li2023seed}. Beyond single-image processing, models such as QwenVL2.5~\cite{qwen2.5-VL} and MiniCPM-o2.6~\cite{yao2024minicpm} can handle multiple images, while video-based LLMs~\cite{zhang2025videollama, maaz:2024video-chatgpt, zhang2024llavanextvideo} preprocess video content by sampling key frames before inference. Additionally, state-of-the-art models like Gemini~\cite{reid2024gemini} and frameworks incorporating interleaved Chain-of-Thought~\cite{gao2024interleaved} explicitly account for the sequential order of both text and images in their reasoning processes. Our approach extends from single to sequential images, mirroring real-world perceptual processes and aligning with emerging trends in multimodal research.


\paragraph{Visual Implicit Meanings Understanding}


Beyond studies on surface-level image understanding~\cite{antol2015vqa, wang2022co, dong-etal-2022-premise, xia-etal-2023-imagenetvc}, recent works have shown that LMMs struggle implicit meaning understanding~\cite{desai2021nice, abu-farha-etal-2022-semeval, Hu2024CrackingTC}.
A recent study~\cite{yang2024:deepsemantic} further highlights a significant gap between AI and human comprehension of implicit meanings in images. However, these works are limited to single-image analysis. Multiple sequential images, arranged temporally, provide richer contextual information and serve as a bridge between static images and videos. Existing studies on sequential images have only focused on surface-level understanding~\cite{chen2024autoeval, wang2024mementos}.
A detailed comparison with prior work is presented in Table~\ref{tab:Features}, and the detailed description of categories and distributions covered by our method are illustrated in Appendix~\ref{categories}.

\section{Dataset and Task Overview}
To systematically evaluate the narrative comprehension capabilities of LMMs on silent comics, we design a suite of tasks arranged in order of increasing complexity. Each task is crafted to probe a distinct facet of visual and sequential reasoning:

\begin{itemize}
\item \textbf{Contextual Frame Prediction:} Assesses the model reasoning ability to predict missing frames in image sequences based contextual cues. Successful performance requires a keen understanding of local visual continuity and narrative coherence.

     \item \textbf{Visual Narrative Comprehension:} Requires the model to infer the overall implicit narrative or “punchline” from multiple frames, moving beyond local cues. 
      
      \item \textbf{Temporal narrative Reordering:} Evaluates whether models correctly infer and restore the chronological order of image sequences based causal temporal relationship. Demands a complete understanding of sequential logic and causal relationships, making it the most challenging of the three.
\end{itemize}

The instructions for three subtasks are presented in Table~\ref{prompt}. These subtasks provide a rigorous and multifaceted assessment of LMMs, offering insights into their strengths and limitations in sequential image understanding. In the following, we may use their full names or refer to them as \textit{prediction}, \textit{comprehension}, and \textit{reordering} for simplicity.

\begin{table}[!ht]
    \centering
    \small
    \begin{tabular}{p{7cm}} 
        \toprule
    \RaggedRight \textbf{Prediction:} \textit{Based on the overall story, what is happening in the blank frame (second-to-last)?} \\ 
    \midrule
        \RaggedRight \textbf{Comprehension:} \textit{What is happening in this comic strip? What is the implicit meaning? Based on your understanding, answer the question.} \\ 
        \midrule
        

           
        \RaggedRight \textbf{Reordering:} \textit{The sequence of the comic strips provided below is incorrect. Your task is to find out the correct order of the comic strips based on the storyline and temporal logical relationship. Number each comic strip in the order they should appear, starting from 1.} \\ 
        \bottomrule
    \end{tabular}
    \caption{Instruction for each tasks.}
    \label{prompt}
    \vspace{-4mm}
\end{table}
\begin{table}[!ht]
\centering
\small
\resizebox{0.48\textwidth}{!}{
\begin{tabular}{ccccc}
\toprule
\textbf{Task} & \#Examples & Length & \#Frames & \#Type \\
\midrule
Prediction & 600 & 19.07 & 2642 & 6 \\
Comprehension & 680 &30.53 & 2762 & 6 \\
Reordering & 890 & 4.08 & 3635 & 6 \\

\bottomrule
\end{tabular}}
\caption{Statistics of \dataset. \#Nums refers to the sum of samples. Length refers to the average of options. \#Frames refers to the sum of frames. \#Type refers to the sum of categories of images.}
\label{tab:statistics}
\vspace{-4mm}
\end{table}


Detailed statistics is displayed on Table~\ref{tab:statistics}.
Overall, our proposed \dataset includes $896$ image sequences, with an average frame length of $4.08$ of each sequence. The number of frames ranges from 3 to 8. Each task is designed in the form of multiple-choice questions, except for the reorder task, which also includes a question-answering format. Since its options are simple and well-defined, accuracy can be directly computed. In our tasks, the input format uniformly consists of images paired with textual prompts. Specifically, the comprehension task utilizes the whole images without split, the reordering task takes shuffled image sequence as input,  and the frame prediction task involves masking second-to-last frame within the image sequence.
For the frame prediction task, we select the second-to-last frame, as it typically serves as a bridge between the preceding frames and the final frame. The start and end frames are generally more challenging to predict.

\section{Dataset Construction}


We construct our \dataset dataset in a multi-step crowd-sourcing pipeline, including 1) annotator training, 2) data annotation, and 3) cross-check examination. An overall demonstration of our dataset construction pipeline is illustrated in Figure~\ref{fig:construct}. 

\subsection{Image Source}

We use silent comic strips, comic with panels and no dialogue, as our primary data source. As a distinct art form, comic strips often encapsulate complex narratives within concise visual sequences, addressing deeper themes such as social satire, humor, and inspiration. These characteristics make comic strips a particularly challenging medium for evaluating ability of LMMs to understand visual sequences. 
The dataset comprises samples from well-known comics, such as \textit{Father and Son} and \textit{Peanuts}, along with web-scraped images from \texttt{GoComics} \footnote{https://www.gocomics.com/}, \texttt{Google}, and \texttt{Facebook} \footnote{https://www.facebook.com/}. 
Initially, we collected $1,260$ images and then refined the dataset through a filtering process. We conducted a thorough manual inspection to eliminate unclear, toxic, overly simplistic images, along withmulti-panel comics lacking a clear temporal sequence. Moreover, comics with dialogue will also be removed to prevent the model from using OCR to understand the meaning of the comics through text rather than through images. As a result, the final dataset was reduced to 896 images.



\subsection{Phase 1: Data Annotation.}

\paragraph{Annotator Training}
We posted job descriptions on online forums and received over 50 applications from candidates with at least a Bachelor's degree. To ensure dataset quality, we provided training sessions that included online pre-annotation instructions and a qualification test to assess candidates' performance. Only those scoring above $95\%$ were selected. candidates were assigned to one of two groups: annotators or inspectors. Ultimately, we hired $13$ annotators and $7$ inspectors for our data annotation process.
To optimize efficiency and reduce costs, we implement a semi-automated pipeline for \dataset annotation, leveraging \texttt{GPT-4o}\footnote{We use the \texttt{gpt-4o-2024-11-20} version for the data annotation process and subsequent evaluations in this work.}. Specifically, our data annotation process consists of two substeps: \textit{answer creation} and \textit{distractor generation}.

\paragraph{Answer Creation.}
The bottom panel of Figure~\ref{fig:construct} illustrates the process of our answer creation phase. Notably, only the comprehension task and frame prediction task need option annotation. The reordering task only necessitates using a program to randomly shuffle the frame order as the answer order, without the need for manual selection of the correct answer. We adopt an AI-assisted annotation approach in which human annotators refine pre-generated answers instead of creating them from scratch. Initially, we leverage \texttt{GPT-4o, GPT-4o-Mini} \cite{hurst2024gpt40} and Gemini \cite{reid2024gemini} to generate diverse candidate answers. Human annotators then evaluate the image sequence and candidate answers, selecting the most appropriate ground truth. If none of the candidate answers is suitable, annotators are instructed to either refine a specific answer or create a new ground truth from scratch, which is about $28\%$. 

\paragraph{Distractor Generation.}
We use candidates with plausible hallucinations from the previous sub-step as strong distractors.  To ensure diversity in the multiple-choice options, we also prompt \texttt{GPT-4o, GPT-4o-mini} \cite{hurst2024gpt40} to generate intentionally incorrect responses as weak distractors. 
Typically, the responses of \texttt{GPT-4o-mini}  are not very accurate and are mostly used as distractors.
Annotators are instructed to evaluate the quality of these distractors and select top-3 options, refining them if necessary. Finally, each ground truth is paired with three high-quality distractors for evaluation.






\subsection{Phase 2: Cross-Check Examination}
We implement a cross-check examination mechanism to ensure rigorous screening of high-quality annotations. During the data annotation process, hired inspectors review the annotated data and corresponding image sequences. If they encounter low-quality annotations, they have the option to reject them. Each annotation is reviewed by two inspectors. If both inspectors reject the annotation, it is discarded, and the image is returned to the dataset for re-annotation. If an image sequence is rejected in two rounds of annotation, it suggests that this sample is not suitable for the current task (e.g., the meaning of the sample is unclear), and the image is subsequently removed from the task. 


After annotation, both advanced annotators and inspectors, acting as final examiners, review the annotations to ensure they meet the required standards. Each annotation undergoes review by three examiners, who vote on whether to accept the annotated sample. Only the samples that receive a majority vote are approved. To ensure the quality of the examiners' work, we randomly sample 10\% of the annotations for verification. 


\subsection{Data Composition}
It is notably that the reordering task does not require human annotation, as described in the previous process. For this task, we select suitable image sequences based on the criterion that the correct ordering must be unique. To ensure this, we conduct a manual review to verify that each sequence follows a logically unambiguous order. A script is then run to perform the initial splitting of panels within specific comics, followed by a random shuffling of these panels. Human annotators are tasked with verifying the format and quality of the frames to ensure they meet the required standards. These processed image sequences serve as the evaluation data for the reordering task.

The final version of our 32-day annotated \dataset contains 896 items (see Table 2), encompassing three tasks: \textit{visual narrative comprehension}, \textit{contextual frame prediction}, and \textit{temporal narrative reordering}. In each of these tasks, each sample consists of an image sequence paired with a multiple-choice question offering four options. The evaluated LMMs are required to select the option they deem most appropriate from the four. More information and examples of \dataset can be found in Appendix~\ref{annotation}.

\definecolor{MyLightOrange}{RGB}{240,200,192}
\definecolor{MyLightgreen}{RGB}{214,240,141}
\definecolor{MyLightblue}{RGB}{160,216,251}

\definecolor{ori}{RGB}{255,165,0}  
\definecolor{grey}{RGB}{248,248,246}
\begin{table*}[!ht]
    \centering
    \renewcommand{\arraystretch}{1.0} 
    \setlength{\tabcolsep}{3pt} 
    \resizebox{\textwidth}{!}{
    \begin{tabular}{lcccccrcc}
    \toprule    
    \rowcolor{ori!8}
    \textbf{Models} & \textbf{Backbone}&  \textbf{\#Params} & \textbf{I-Video}  &  \textbf{Prediction} & \textbf{Comprehension} &\textbf{ R-C} & \textbf{R-G } & \textbf{AVG} \\ 
   
    \midrule
    
    \rowcolor{grey!2}
    
    Human  & &  & & 82.00 & 80.00 & 86.00 & 80.00 & 82.00\\ 
    \midrule
    \rowcolor{MyLightblue!30}
    \multicolumn{9}{c}{\textbf{Closed-Model}} \\
   \midrule
 \rowcolor{MyLightblue!5}
    GPT-4o-mini \cite{hurst2024gpt40} & - &- & & 56.33 & 53.23 & \textbf{26.07} & 8.45 & 36.02 \\
    \rowcolor{MyLightblue!5}
    Gemini1.5-Pro \cite{reid2024gemini}& - & -& & 67.83 & 49.56  & 25.51 & \textbf{32.02 } & 43.73 \\ 
    \rowcolor{MyLightblue!5}
    GPT-4o \cite{hurst2024gpt40} & - &-  & &\textbf{69.95} & \textbf{61.60} & 25.17 & 23.93 & \textbf{45.16} \\ 
    
    \midrule
    
   \rowcolor{MyLightgreen!30}
    \multicolumn{9}{c}{\textbf{Open-Source}} \\
    \midrule  
    \rowcolor{MyLightgreen!5}
    Janus-Pro \cite{chen2025januspro} & DeepSeek-LLM-7b-base & 7B &\ding{55} & 27.50 & 27.50 & 26.07 & *    & 20.27 \\
    \rowcolor{MyLightgreen!5}
    mPlug-Owl2 \cite{ye2023mplugowl2} &   LLaMA2            & 8B &\ding{55}  & 31.17 & 30.74 & 25.06 & 0.56 & 21.88\\
    \rowcolor{MyLightgreen!5}
    LLaVA-v1.6 \cite{liu2023llava}  & Vicuna-v1.5           & 7B &\ding{55}  & 43.50 & 34.41 & 26.29 & 3.37 & 26.89 \\
    \rowcolor{MyLightgreen!5}
    LLaVA-NeXT-Video~\cite{zhang2024llavanextvideo}  & Vicuna-v1.5  & 7B &\ding{51} & 44.50 & 45.74 & 23.71 & *    & 28.49 \\
    \rowcolor{MyLightgreen!5}
    CogVLM \cite{wang2023cogvlm} & Vicuna-v1.5              & 17B &\ding{55} & 56.00 & 34.26 & 24.83 & *    & 28.77 \\
    \rowcolor{MyLightgreen!5}
    LLaVA-v1.6 \cite{liu2023llava}  & Vicuna-v1.5           & 13B&\ding{55}  & 46.50 & 46.03 & 27.98 & 2.58 & 30.77 \\
    \rowcolor{MyLightgreen!5}
    LLaVA-v1.6 \cite{liu2023llava}  & Vicuna-v1.5           & 34B &\ding{55} & 50.83 & 52.94 & 25.62 & 2.13 & 32.88 \\
    \rowcolor{MyLightgreen!5}
    Cambrian \cite{tong2024cambrian1}  & Vicuna-v1.5        & 13B &\ding{55} & 55.00 & 45.59 & 26.85 & 4.94 & 33.10\\
    \rowcolor{MyLightgreen!5}
    Qwen2.5VL \cite{qwen2.5-VL}      & Qwen2.5              & 3B  &\ding{51} & 58.83 & 50.59 & 27.75 & 1.91 & 34.77 \\
    \rowcolor{MyLightgreen!5}
    InternVL2v5 \cite{chen2024internvl} &   Intern          & 26B &\ding{51}  & 65.17 & 60.92 & 24.61 & 2.58 & 38.32\\
    \rowcolor{MyLightgreen!5}
    MiniCPM-o 2.6  \cite{yao2024minicpm} & Qwen2.5          & 7B  & \ding{51} & \textbf{65.83} & 56.18 & 26.85 & 5.51 & 38.59\\ 
    \rowcolor{MyLightgreen!5}
    Qwen2.5VL \cite{qwen2.5-VL} & Qwen2.5                   & 7B &\ding{51}  & 64.00 & 56.03 & 29.21 &\textbf{ 11.01} & 40.06\\ 
    \rowcolor{MyLightgreen!5}
    Qwen2VL \cite{Qwen2VL} & Qwen2                          & 7B &\ding{51} & 63.00 & \textbf{58.53} & \textbf{31.91} & 9.44 & \textbf{40.72}\\
   
    \bottomrule
    \end{tabular}}
    \caption{Comparison of model performance across various architectures and scales, sorted by accuracy. I-Video indicates support for video input. R-C and R-G represent the Reordering task with choice and generation formats, respectively. Scores are reported as percentages (\%), with * denoting a model failure on the corresponding task. AVG is the average of all four scores, including failures. Bold values highlight the highest scores among closed-source and open-source models.}
    \label{tab:model_comparison}
    \vspace{-4mm}
\end{table*}

\section{Experiments}

\subsection{Models}
To comprehensively evaluate on LMMs, we conducted zero-shot inference across both commercial and open-source models. Our evaluation suite includes leading commercial models GPT-4o~\cite{hurst2024gpt40} and Gemini1.5-Pro~\cite{Gemini} alongside state-of-the-art open-source alternatives of varying scales: Qwen2.5-VL~\cite{qwen2.5-VL}, Qwen2-VL~\cite{wang2024qwen2}, LLaVA-v1.6~\cite{liu2023llava}, CogVLM~\cite{wang2023cogvlm}, MiniCPM-o-2.6~\cite{yao2024minicpm}, mPlug-Owl2~\cite{ye2023mplugowl2}, InternVL2v5~\cite{chen2024internvl},LLaVA-NEXT-Video~\cite{zhang2024llavanextvideo} and Cambrian~\cite{tong2024cambrian1}. Besides, Janus-Pro~\cite{chen2025januspro}, which unifies multimodal understanding and generation, is included to test the abilities between Unified Model and Vision Language Model.  This diverse selection enables us to analyze how model scale, architecture, and training approaches influence comic comprehensive capabilities.


\subsection{Experimental Details}
The task prompts is displayed in Table ~\ref{prompt}. For visual narrative comprehension task, model is provided with the whole image. But for next-frame prediction and multi-frame sequence reordering task, LMMs infer with image sequences.
The hyper-parameters for each LMMs in the experiments including possible settings are detailed in Appendix~\ref{appendix:hyper-param}. Furthermore, to assess human capabilities in these tasks, we randomly select 100 questions from the dataset for each task and instruct human evaluators to answer. This allows us to benchmark the performance of human participants against our models, offering a thorough comparison of both human and LMMs proficiency in these specific tasks. We present sample outputs of three tasks generated by LMMs in Figure~\ref{fig:case}.

\subsection{Main Results}
Our comprehensive evaluation reveals that while LMMs show promising capabilities in comprehension and prediction tasks, they significantly underperformed in sequence reordering tasks. Moreover, there remains a substantial performance gap between current models and human performance across all tasks. Unified Model underperformed than Vision Language Model.

\paragraph{Contextual Frame Prediction}
The frame prediction task appears to be the most tractable among the three tasks. GPT-4o achieves the highest score of $69.95\%$, followed closely by Qwen2-VL at $64.00\%$.This demonstrates that the performance gap between closed and open-source models is relatively small for this task. However, Janus-Pro perform notably below expectations ($27.50\%$), possibly due to its unified model architectural.

\paragraph{Visual Narrative Comprehension}
For visual narrative comprehension, we observe a similar pattern but with generally lower scores. GPT-4o leads with $61.60\%$, while other models show varying degrees of capability. 

\paragraph{Temporal Narrative Reordering}
The frame reordering task proves to be the most challenging, with all models performing significantly below human capability. Even the best-performing models struggle to exceed $30\%$ accuracy, with many achieving scores around $25-26\%$, which is slightly higher than random selection. 
Notably, several models (marked with *) are unable to perform this task due to their architectural limitations in processing multiple images simultaneously. For these models, we attempted to accommodate their single-image constraint by concatenating multiple frames horizontally into a single image, with white margins serving as frame boundaries. However, this workaround appears to be suboptimal, as these models likely struggle to properly distinguish individual frame boundaries and maintain the semantic independence of each frame, ultimately leading to their poor performance on the reordering task. 

The poor performance on reordering task suggests that current LMMs, regardless of their scale or architecture, have not yet developed robust capabilities for understanding temporal relationships and sequential logic in visual narratives.

\section{Analysis}
Our analysis addresses the following questions:
\begin{figure*}[t]
\centering
\includegraphics[width=0.99\textwidth]{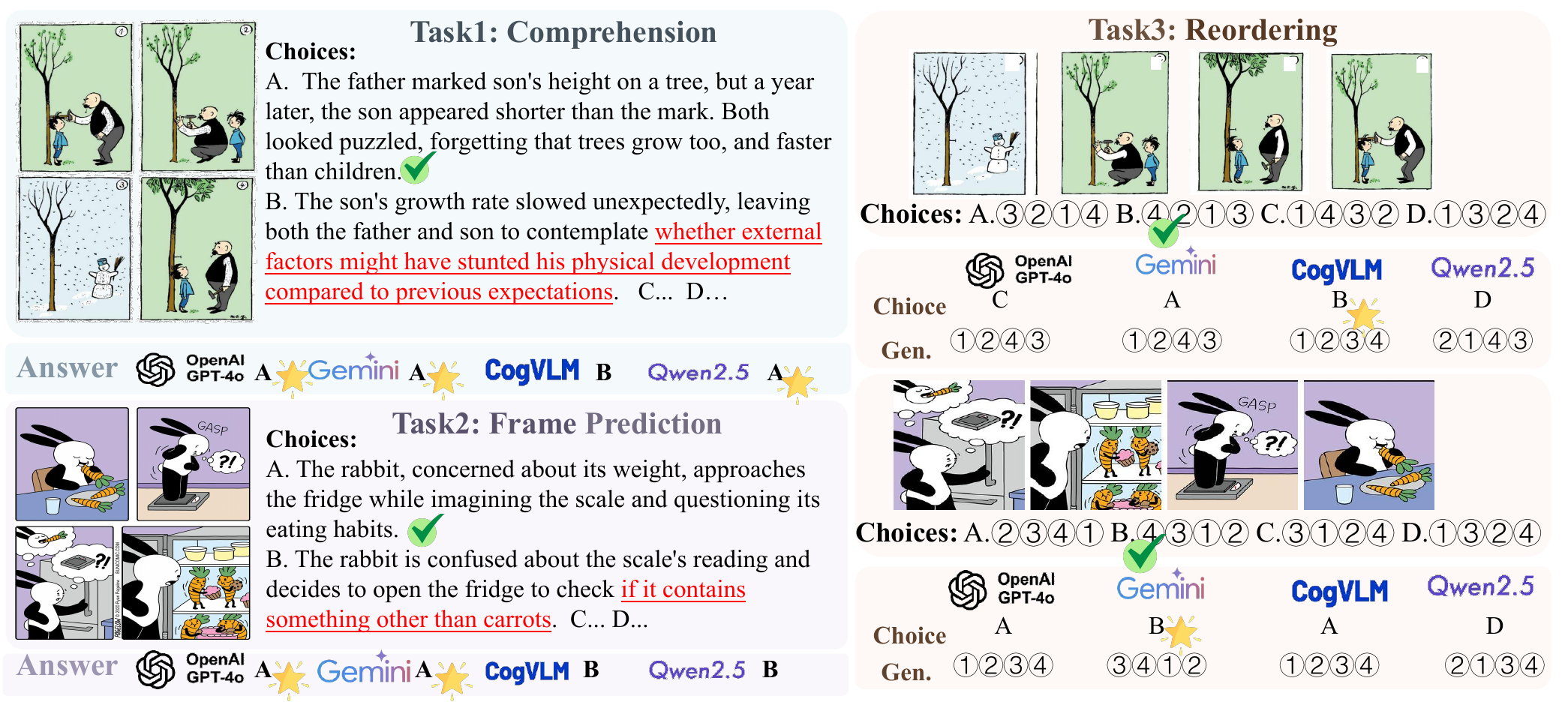}
\caption{Sample outputs of our three tasks generated by different vision language models, along with gold truth. We highlight errors in distractors. }
\label{fig:case}
\vspace{-4mm}
\end{figure*}

\paragraph{Does fine-tuning with reorder task help?}
Yes, it does. We fine-tune Qwen2-VL using 3,160 samples for one epoch. This not only significantly improves performance on the reordering VQA task but also enhances comprehension tasks.
To construct the training dataset, we applied data augmentation to 790 images using the reorder task. Specifically, we randomly shuffled the sequence of images four times, generating a total of approximately 3,160 distinct samples. For evaluation on the reorder task, we used only the remaining 100 samples. For the comprehension task, we conducted a full test set evaluation, as the training data provided only images without any analytical content. Meanwhile, we excluded the frame prediction task from testing due to potential data leakage. The experimental results are presented in the table~\ref{tab:finetune}. 
Overall, our reordering data is useful for fine-tuning, as it can enhance the LMMs to reason on sequential images. However, the ultimate performance still depends on the base capability of the model. According to the table, Qwen2.5-VL significantly benefits from fine-tuning, enhancing both reordering and comprehension tasks. Improvements in generation tasks for reordering are greater than in choice tasks, likely because generation instructions are more challenging and unfamiliar, while choice tasks allow educated guesses from provided options. Limited training samples also constrain improvements in choice tasks. In contrast, LLaVA only improves in the reorder-choice task after training.

\begin{table}[htbp]
    \centering
     \resizebox{0.45\textwidth}{!}{
    \begin{tabular}{c|c|c|c}
    \toprule
        Tasks  & Comprehension & Reordering-G & Reordering-C \\
        \midrule
       Qwen2-VL & 58.53 & 6.00 & 31.00 \\
       +finetune & 62.94 & 31.00 & 38.00 \\
       \midrule
       LLaVA-1.6 & 34.41 &3.00& 26.00 \\
       +finetune & 33.82 &2.00 & 32.00 \\
       \bottomrule
    \end{tabular}}
    \caption{Performance on Qwen2-VL-7B and LLaVA-1.6-7B finetuned with reordering task data. Reordering-C refers to the Reordering-choice. Reordering-G refers to the Reordering-generation}
    \label{tab:finetune}
    \vspace{-5mm}
\end{table}

\paragraph{Does GPT-4o understand sequence images as well as humans?}
While GPT-4o achieves the highest performance among all tested models, there remains a substantial gap between its capabilities and human performance, particularly in novel tasks like frame reordering. Through our preliminary data annotation experiments, we observed that while GPT-4o can comprehend basic comic content and provide interpretations, it frequently generates hallucinated content and struggles with comics that depict unconventional or imaginative scenarios rarely encountered in real life.

In the visual narrative comprehension and next-frame prediction tasks, the multiple-choice format allows models to leverage similarity matching between options. Our investigation revealed that model performance is heavily influenced by the quality of distractor options. In initial experiments with weak distractors (generated using GPT-4o with instructions to provide distractors with hallucinations, the prompt is followed HalluEval), the model achieved accuracy rates up to 90\%. Upon analysis, we found these initial distractors were too obviously incorrect or irrelevant to the comic content, making the selection task trivial. To address this limitation, we carefully curated a new set of challenging distractors. With these enhanced distractors, performance of GPT-4o decreased significantly to more realistic levels ($61.60\%$ for understanding and $69.95\%$ for prediction), better reflecting the true challenges in comic comprehension. The scores obtained from multiple-choice questions with semantically transparent options tend to be inflated. In subsequent reordering tasks, where options lack explicit semantic meanings, coupled with open-ended questions, the scores provided a more authentic assessment of the LMMs.

\paragraph{Does input format of images influence performance?}

Considering the distinct computational pathways that LMMs employ in processing individual versus multiple images, we designed following experiments to measure the differential impact of varied input formats using Qwen2.5VL as our test case. We compared three input formats: (1) whole image - the entire comic strip as a single image, (2) sequential frames - individually separated frames input in order, and (3) shuffled sequence - separated frames input in random order. Table~\ref{tab:understanding} shows surprisingly consistent performance across all three formats ($56.03\%$, $54.56\%$, and $57.65\%$ respectively).

This consistency suggests that while separated frames might theoretically help models extract clearer information from each panel and avoid visual confusion from complex layouts, the current video-like processing mechanism used by LMMs for multiple images might not fully capitalize on these advantages. The similar performance with shuffled sequences further indicates that models rely more on individual frame content rather than sequential relationships for comprehension tasks.

\begin{table}[htbp]
    \centering
     \resizebox{0.45\textwidth}{!}{
    \begin{tabular}{c|c|c}
     \toprule
       Input  & GPT-4o-mini & Qwen2.5-VL \\
       \midrule
       Whole Image  & 53.23 & 56.03 \\
      Image Sequence & 51.03 &  54.56 \\
       Shuffled Sequence & 49.56 & 57.65 \\
       \bottomrule
    \end{tabular}}
    \caption{Performance with different input format for understanding task. }
    \label{tab:understanding}
    \vspace{-6mm}
\end{table}
\paragraph{Does implicit meaning help reordering task?}
To investigate whether poor reordering performance stems from inadequate semantic understanding, we enhanced the reordering task by providing explicit semantic annotations along with shuffled images\footnote{The prompt is displayed in Figure~\ref{fig:prompt-meaning}}. As shown in Table~\ref{tab:reorder_enhanced}, this additional semantic information only marginally improved performance (from 30.01\% to 32.54\%). This modest improvement suggests that the bottleneck in reordering tasks lies not in semantic understanding but in the fundamental capability to reason about temporal and logical sequences in visual narratives.

\begin{table}[htbp]
    \centering
     \resizebox{0.45\textwidth}{!}{
    \begin{tabular}{c|c|c}
    \toprule
        Input & GPT-4o-mini & Qwen2.5-VL \\
        \midrule
       Shuffled image & 26.07  & 30.01 \\
       +Meaning & 28.40 & 32.54 \\
       \bottomrule
    \end{tabular}}
    \caption{Performance with enhanced data (correct answer for comprehension task) for reordering task.}
    \label{tab:reorder_enhanced}
    \vspace{-4mm}
\end{table}



\begin{figure}[t]
\centering
\includegraphics[width=0.96\columnwidth]{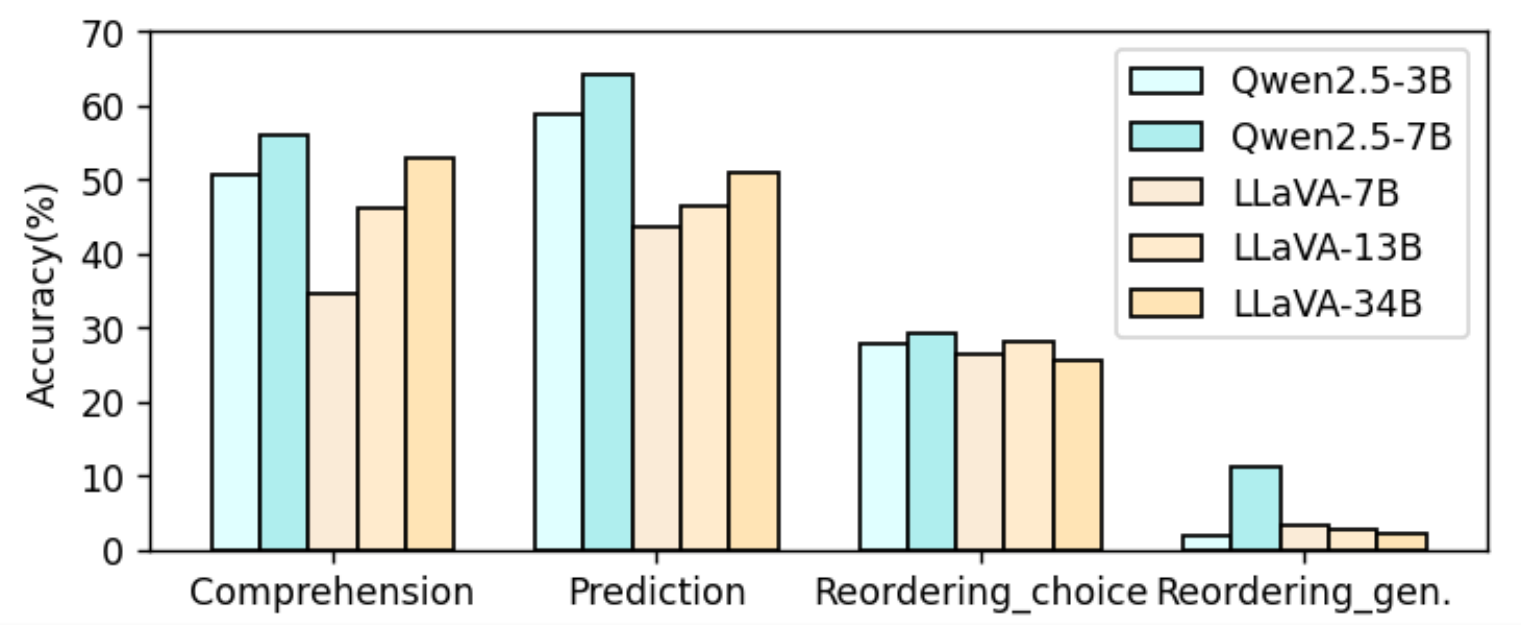}
\caption{ Comparison of the accuracy results between Qwen2.5-3B vs Qwen2.5-7B and LLaVA-1.6-7B vs LLaVA-1.6-13B vs LLaVA-1.6-34B}
\label{fig:scale}
\vspace{-5mm}
\end{figure}

\paragraph{How does model size affect?}
In this section, we will discuss the relationship between model parameters size and reasoning performance for tasks due to the scaling law. We examine on LLaVA and Qwen2.5VL, from 3B scale to 34B scale.
There is a clear scaling effect across model sizes, as demonstrated by LLaVA1.5's performance improving from $34.41\%$ (7B) to $46.03\%$ (13B) to $52.94\%$ (34B). This suggests that model scale plays a crucial role in comprehending implicit meanings in visual narratives.
Figure~\ref{fig:scale} provide a visual representation of the performance trend for five models.

\paragraph{How does each task influence others}
Table~\ref{tab:finetune} demonstrates that fine-tuning with reordering data enhances comprehension performance, likely because it compels the model to capture subtle narrative transitions. Table~\ref{tab:understanding} confirms the importance of correct temporal order, as shuffling image sequences significantly degrades comprehension performance. Table~\ref{tab:reorder_enhanced} reveals that models with robust comprehension skills tend to reorder frames more accurately, highlighting the synergistic relationship between narrative understanding and reordering capability.




We also analyzed additional factors, including comic category and frame count effects on model inference performance. These detailed analyses are provided in Appendix~\ref{appendix:analysis} due to space limitations.

\section{Conclusion}

We present \dataset{}, a comprehensive benchmark for evaluating Large Multimodal Models' capabilities in visual comic sequence reasoning. Our benchmark comprises meticulously curated and human-AI annotated tasks spanning visual narrative comprehension, next-frame prediction, and multi-frame sequence reordering. Through extensive evaluations of state-of-the-art LMMs, we identify significant performance gaps between AI systems and human capabilities in comic strip understanding.
These findings underscore the considerable challenges that remain in developing AI systems capable of deep visual semantic understanding comparable to human cognition. 

\section*{Limitations}
\label{subsec:limitation} 

\textbf{Limited Availability of Comic Strips:} Our dataset contains a relatively small number of samples due to the scarcity of standalone short-story comic strips available online. Most comics are either serialized narratives or dialog-driven, making it challenging to collect a diverse set of independent stories.


\textbf{Limited Training Data for Fine-Tuning:} Our findings indicate that fine-tuning significantly enhances model performance on the reordering task. However, the limited availability of training data constrains the model’s ability to fully develop temporal reasoning skills. Expanding the dataset or incorporating alternative sources, such as video sequences, could further improve performance.

\section*{Ethics Statement}
\label{subsec:ethics} 
The datasets used in our experiment are publicly released and labeled through interaction with humans in English. In this process, user privacy is protected, and no personal information is contained in the dataset. The scientific artifacts that we used are available for research with permissive licenses. And the use of these artifacts in this paper is consistent with their intended use. Therefore, we believe that our research work meets the ethics of ACL. 

\section*{Acknowledgments}
\label{subsec:ack} 
This paper is sponsored by State Key Laboratory of Multimedia Information Processing Open Fund.

\bibliography{custom}

\clearpage

\appendix

\section*{Appendix}

\section{License and Copyright.}
We used original web links to comic images without infringing on their copyright. This work is licensed under a CC BY-NC license. We will open-source all related code for processing image sequences and frames to facilitate the reproducibility of our evaluated image sequences. All annotators participated voluntarily in the annotation process and were provided fair compensation.

\section{Model}
\label{appendix:model}
Our evaluation suite includes leading commercial models GPT-4o~\cite{hurst2024gpt40} and Gemini1.5-Pro~\cite{Gemini} alongside state-of-the-art open-source alternatives of varying scales: Qwen2.5-VL~\cite{qwen2.5-VL}, Qwen2-VL~\cite{wang2024qwen2}, LLaVA-v1.6~\cite{liu2023llava}, CogVLM~\cite{wang2023cogvlm}, MiniCPM-o 2.6~\cite{yao2024minicpm}, mPlug-Owl2~\cite{ye2023mplugowl2}, InternVL2v5~\cite{chen2024internvl},LLaVA-NEXT-Video~\cite{zhang2024llavanextvideo} and Cambrian~\cite{tong2024cambrian1}. Besides, Janus-Pro~\cite{chen2025januspro}, which unifies multimodal understanding and generation, is included to test the abilities between Unified Model and Vision Language Model.

\section{Model Hyper-parameter Details}
\label{appendix:hyper-param}

We use the default hyper-parameter values of the models. In the LLaVa-1.5-7B and LLaVa-1.5-13B, the temperature is set to 0.2. For MiniGPT-4, the temperature is set to 1.0, and num\_beams is also set to 1.0. The temperature for mPlug-Owl-2 is set to 0.7. For CogVLM, the temperature is set to 0.4, top\_p is set to 0.8, and top\_k is set to 1.0.

In the LLaVa-1.6-7B, LLaVa-1.6-13B and LLaVa-1.6-34B, the temperature is set to 0.2. In the Qwen2-VL-7B, Qwen2.5-VL-3B and Qwen2.5-VL-7B, the temperature is set to 0.01, top\_p is set to 0.001, and top\_k is set to 1. For CogVLM-17B, the temperature is set to 0.4, top\_p is set to 0.8, and top\_k is set to 1.0. For InternVL2-26B, do\_sample is set to False. For Cambrian-13B, the temperature is set to 0.2.


\section{Analysis}
\label{appendix:analysis}
\paragraph{Where do LMMs fail?}
We present sample outputs of three tasks generated by vision language models (VLMs) in Figure~\ref{fig:case}. These images are easy for human but hard for VLMs. VLMs can understand one comic strip but they can still make mistakes with reordering task. 

\paragraph{Frame counts}
We static the model performance on comprehension task on different frame numbers, as shown in table~\ref{frame}.
Our results indicate that models perform better on both four-panel and six-panel comics. This performance boost is likely due to the uniform dimensions and layouts, which facilitate the reliable identification and sequencing of sub-images.

\begin{table}[!ht]
    \centering
     \resizebox{0.48\textwidth}{!}{
    \begin{tabular}{lllll}
    \hline
        \textbf{Model} & 3 Frames & 4 Frames & 5 Frames & 6 Frames  \\ \hline
        \textbf{Qwen2VL-7B} & 57.06 & 60.51 & 53.09 & 55.26  \\ 
        \textbf{Qwen2.5VL-7B} & 52.76 & 58.99 & 48.15 & 63.16  \\ 
        \textbf{LLaVAv1.6-4B} & 52.76 & 53.16 & 50.62 & 57.89  \\ 
        \textbf{LLaVAv1.6-13B} & 42.94 & 46.84 & 48.15 & 47.37  \\ 
        \textbf{LLaVAv1.6-7B} & 34.97 & 32.41 & 41.98 & 39.47  \\ 
        \textbf{CogVLM-7B} & 33.74 & 33.92 & 35.8 & 36.84  \\ 
        \textbf{AVG} & 45.71 & 47.64 & 46.30 & 50.00 \\ \hline
    \end{tabular}
    }
    \caption{Comprehension task performance comparision of different LLMs on different frames numbers.}
    \label{frame}
\end{table}

\paragraph{Category}
\label{categories}
We evaluated the performance of LLMs on comprehension tasks across 6 categories of comic strip. 

Consistency Across Models: Models such as Gemini1.5-Pro, Qwen2.5VL-7B, and Qwen2VL-7B demonstrate more balanced performance across various categories compared to the higher variability observed with LLaVA-34B, highlighting the robustness of their training strategies.

\begin{figure}[h]
    \centering
    \includegraphics[width=\linewidth]{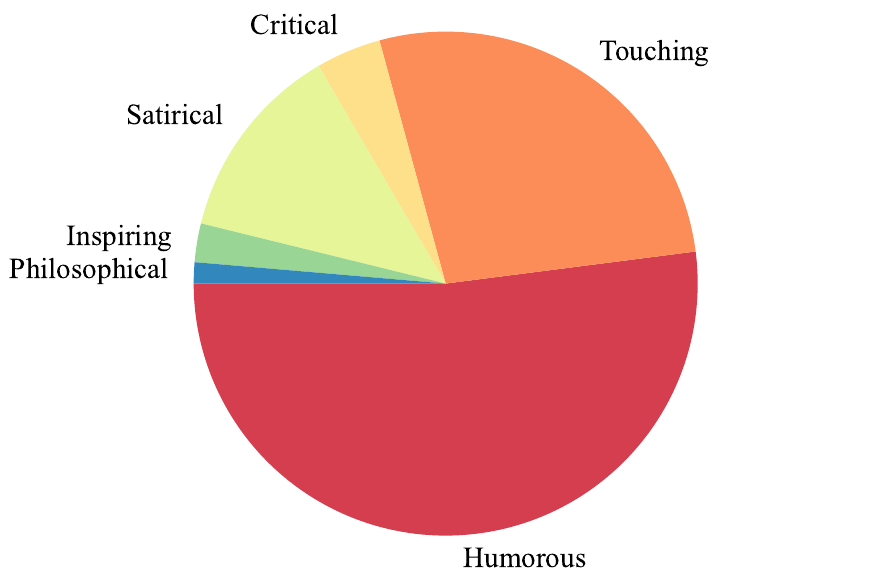}
    \caption{The distribution of six categories of \dataset.}
    \label{fig:pie}
\end{figure}

\begin{figure}[h]
    \centering
    \includegraphics[width=\linewidth]{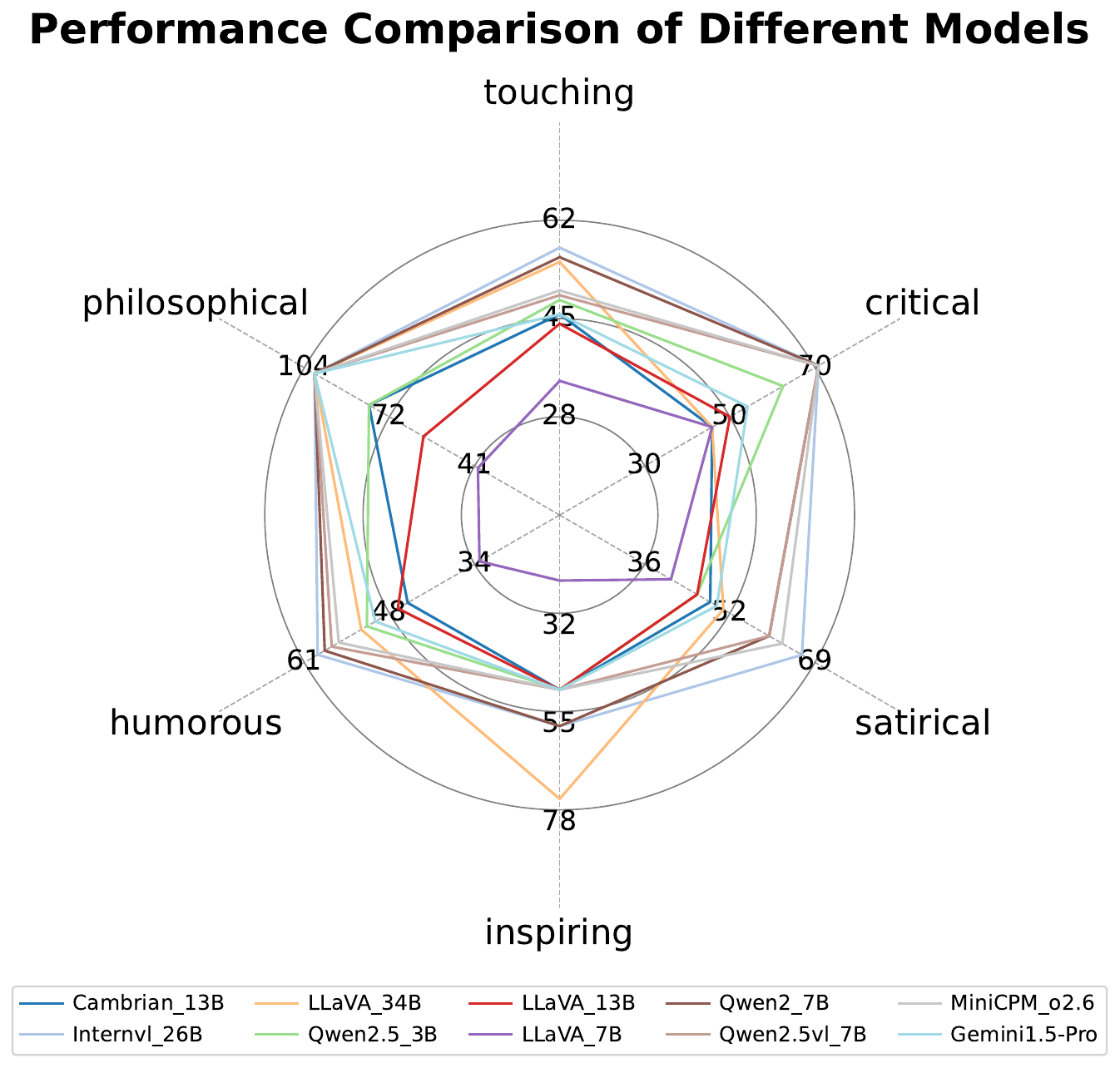}
    \caption{Comprehension task performance comparision of different LLMs on different categories.}
    \label{fig:radar}
\end{figure}

\paragraph{Error Analysis}
During the GPT-4o annotation stage of dataset construction, we also observed instances of hallucinated content. For example, models sometimes generated actions that did not occur or confused relationships between characters (e.g., misidentifying kinship roles or misinterpreting narrative cues). In some cases, comics require external common sense for accurate interpretation. An example of this is analyzing the meaning of the scene in which "eight planets are celebrating around the Sun, while one small planet remains isolated" after Pluto was removed from the list of the nine major planets in the solar system in 2006.
\begin{figure*}
    \centering
    \includegraphics[width=0.8\linewidth]{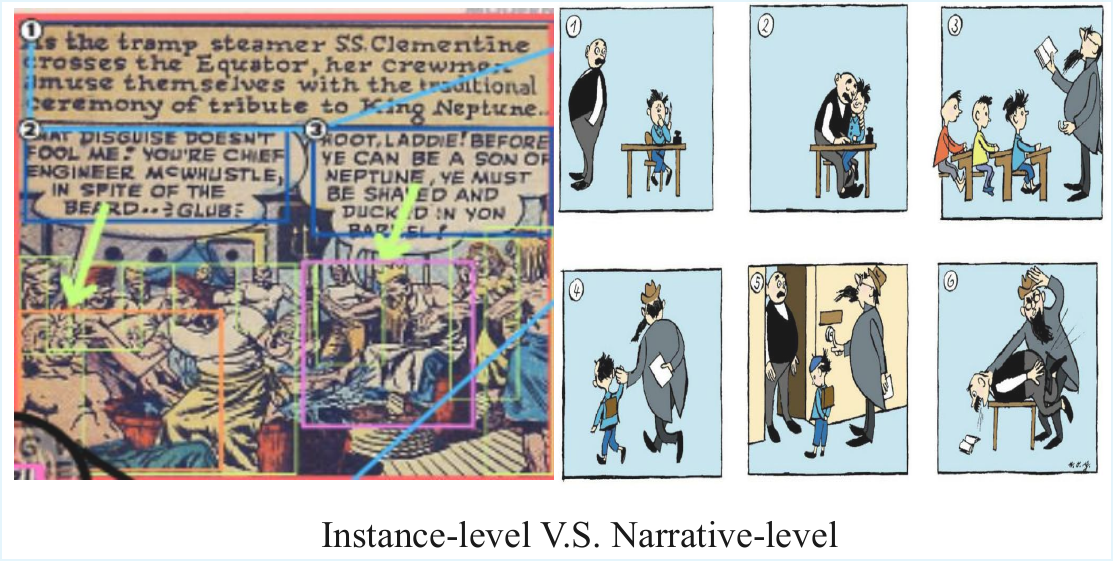}
    \caption{Example for our task and traditional task.}
    \label{fig:compare-task}
\end{figure*}

\begin{figure*}
    \centering
    \includegraphics[width=0.9\linewidth]{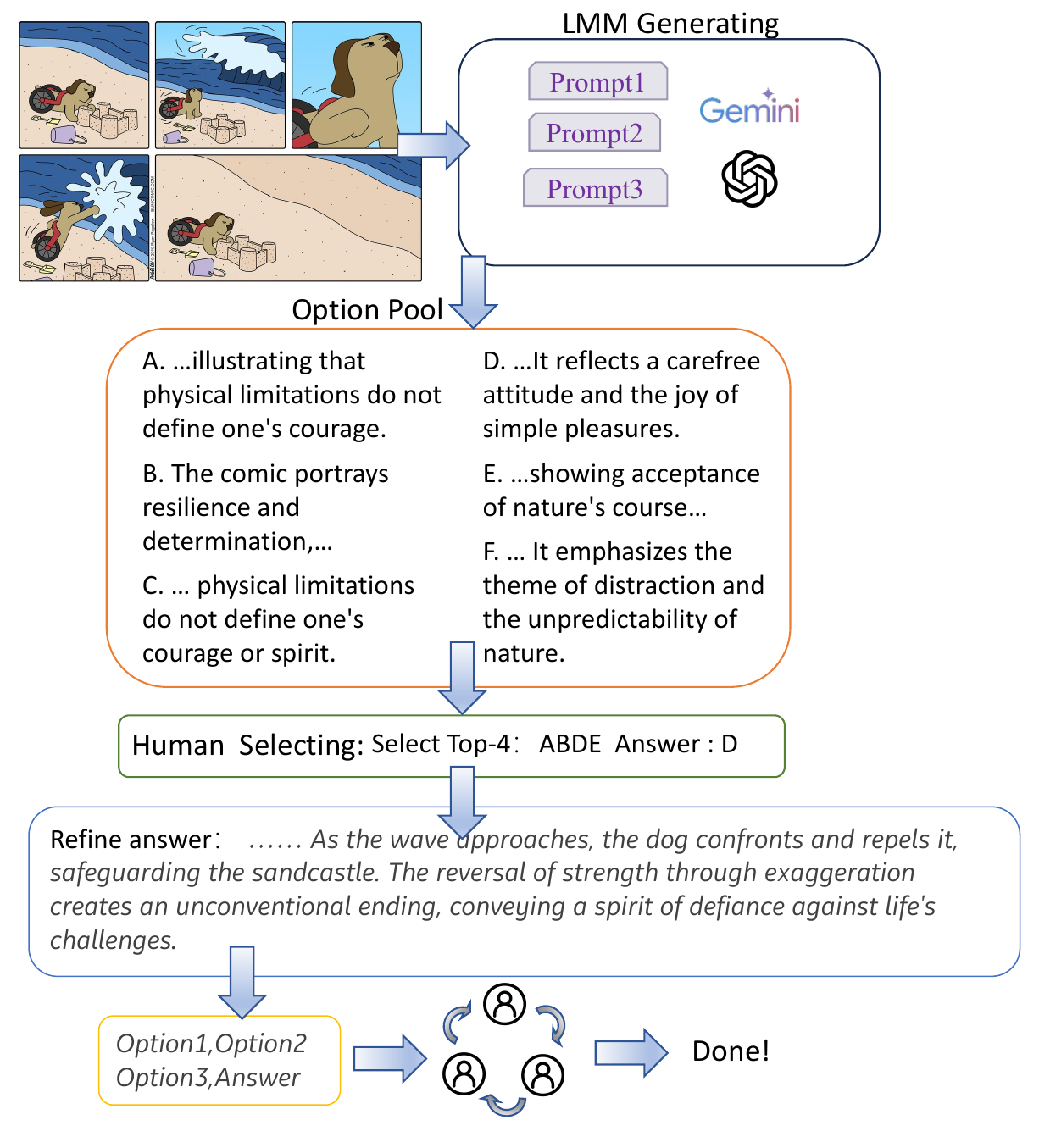}
    \caption{An detailed example of data construction.}
    \label{fig:examples}
\end{figure*}

\section{Examples for comic task comparesion}
Here is an example~\ref{fig:compare-task} of our dataset and CoMix dataset for comparison.

 \section{Examples on data construction}
\label{annotation}
The following listed prompts are used to construct data. By instructing to different LMMs, we can obtain option candidate pool. Here is a detailed example at Figure~\ref{fig:examples}.

\begin{table*}[]
    \centering
    \begin{tabular}{lp{13cm}}
    \toprule
       \textbf{Category}  &  \textbf{Definition} \\ 
       \midrule
         Satirical &  \RaggedRight The comic uses irony, exaggeration, or ridicule to criticize social, political, or cultural issues. It often highlights contradictions, hypocrisy, or absurdity in a way that provokes thought or debate. The humor may be sharp or biting but serves a critical purpose. \\ 
         \midrule
Inspiring& \RaggedRight The comic presents a positive or uplifting message, often encouraging personal growth, motivation, or perseverance. It may depict acts of kindness, success against adversity, or wisdom that encourages the reader to strive for betterment.\\ \midrule
Touching& \RaggedRight The comic evokes emotions such as empathy, nostalgia, or affection. It may explore themes of love, friendship, loss, or family bonds, aiming to create a sentimental or heartfelt response from the audience.\\ 
\midrule
Philosophical& \RaggedRight The comic explores deep, abstract, or existential ideas about life, morality, meaning, or human nature. It prompts the reader to reflect on profound questions, often using metaphors or thought-provoking dialogue rather than direct humor or emotion.\\
\midrule
Critical& \RaggedRight The comic highlights flaws or problems in society, institutions, or human behavior with a serious or analytical tone. Unlike satire, which uses humor as a tool for critique, a critical comic may adopt a more straightforward, serious, or thought-provoking approach to expose issues and encourage awareness.\\ 
\midrule
Humorous& \RaggedRight The comic's primary goal is to entertain and amuse the audience. It relies on lighthearted jokes, wordplay, or visual gags without necessarily conveying a deeper message or critique. The tone is playful, aiming for laughter rather than serious reflection. \\
\bottomrule
    \end{tabular}
    \caption{The types and definition of the categories in \dataset.}
    \label{tab:categories}
\end{table*}

\begin{figure*}[th!]
\begin{AIBox}{}

\parbox[t]{1.0\textwidth}{{\textbf{Prompt1 for prediction task:}} 
\small\begin{alltt}
Predict what happened in the blank panel? Output in 35 words.
\end{alltt}}
\tcbline

\parbox[t]{1.0\textwidth}{{\textbf{Prompt1 for prediction task:}} 
\small\begin{alltt}
You are now a mature hallucination generator. Please generate one strong distractor option for the following question. You can use any method you have learned that is misleading for the given question. \\
Question: Predict what happened in the blank panel. Please output with 35 words without any additional text or explanation.
\end{alltt}}

\parbox[t]{1.0\textwidth}{{\textbf{Prompt1 for comprehension task:}} 
\small\begin{alltt}
What happened in comic strip? Conclude the whole story, then carefully analyze the implicit meaning of comic. Ouput in 35 words.

\end{alltt}}
\tcbline

\parbox[t]{1.0\textwidth}{{\textbf{Prompt2 for comprehension task:}} 
\small\begin{alltt}
You are now a mature hallucination generator. Please generate one strong distractor option for the following question. You can use any method you have learned that is misleading for the given question.
\end{alltt}}

\end{AIBox}

\caption{Instruction for annotating dataset with LLM.}
\label{fig:construct_prompt1}
\end{figure*}

\begin{figure*}[th!]
\begin{AIBox}{}

\parbox[t]{1.0\textwidth}{{\textbf{Prompt for analysis experiment:}} 
\small\begin{alltt}
The sequence of the comic strips provided below is incorrect. Your task is to determine the correct order based on the storyline, temporal relationships, and common-sense logic. **Here is the depiction of the comic strips: '{qa.get('understanding')}'**

Number each comic strip in the order they should appear, starting from 1.

[Options]

Please output the correct option ID without any additional text or explanation
\end{alltt}}
\end{AIBox}

\caption{Instruction for providing implicit meaning on reordering task.}
\label{fig:prompt-meaning}
\end{figure*}

\begin{figure*}[th!]
\begin{AIBox}{}

\parbox[t]{1.0\textwidth}{{\textbf{Prompt3 for comprehension task:}} 
\small\begin{alltt}
 Task Overview:\\
Strive to understand this story and analyze its implicit meaning, then complete multi-choice question for test. You should act in two roles to complete tasks. \\
Image Context:\\
Pic1: The first picture shows the complete comic strip. Read it from left to right, top to bottom, to understand the full narrative arc.\\
Pic2: The second picture is the second-to-last frame from Pic1, which is the target frame in Task 1.\\
Role 1 - Excellent Comic Analysis Expert:\\
Task 1: Contextual Scene Description\\
Question 1: Based on the overall story, what is happening in the second-to-last frame (Pic2) of the comic strip?\\
Requirements:\\
1. Provide a clear and detailed description of the key visual elements, characters, relationships and actions in Pic2. 2. Ensure narrative continuity with the events of the entire comic. 3. Output this as the right option for Task 2 with 30-40 words.\\
Task 2: Implicit Meaning Analysis\\
Question 2: What happened in comic strip (Pic1)? Describe the whole story in detail, then analyze its implicit meaning.\\
Requirements:\\
1. Describe the whole story in detail and analyze its implicit meaning and sentiment. 2. Provide three sentences with 40-50 words as right option for Task 4.\\
Role 2 - Strong Distractor Options Generator:\\
Task 3: Frame Scene Options Generation\\
Question 1: Based on the overall story, what is happening in the second-to-last frame (Pic2) of the comic strip?\\
Requirements:\\
1. Generate three plausible but incorrect options for Question 1. 2.The length of each option should be similar with the correct answer from Task 1 (around 30-40 words)! 3. Ensure that the incorrect options are consistent with the overall story but misinterpret the events of Pic2. \\
Task 4: Comic Strip Analysis Options Generation\\
Question 2: What happened in comic strip (Pic1)? Describe the whole story in detail, then analyze its implicit meaning.  \\
Requirements:\\
1. Generate three plausible but incorrect options for Question 2. 2. Each incorrect option should be composed of 3 sentences and share the same length with the correct answer from Task 2! 3. Avoid obviously wrong or nonsensical answers.
\\
Please strictly adhere to the word count requirement! The final output should be in the following format:\\
Question 1: Based on the overall story, what is happening in the second-to-last frame (Pic2) of the comic strip?\\
Reasoning Chain 1: [Describe the story in sequence.]\\
Options in Q1: \\
A. [Hallucination option with 30-40 words from Task 3] \\
B. [Hallucination option with 30-40 words from Task 3] \\
C. [Hallucination option with 30-40 words from Task 3] \\
D. [Right option with 30-40 words from Task 1] \\
Right Answer 1: D

Question 2: What happened in comic strip (Pic1)? Describe the whole story in detail. Carefully analyze the implicit meaning of comic.\\
Reasoning Chain 2: Reason step by step here.\\
Options in Q2: \\
 A. [Hallucination option with 40-50 words from Task 4]\\
B. [Hallucination option with 40-50 words from Task 4] \\
C. [Hallucination option with 40-50 words from Task 4] \\
D. [Right option with 40-50 words from Task 2] \\
Right Answer 2: D 
\end{alltt}}

\end{AIBox}

\caption{Instruction for annotating dataset with LLM.}
\label{fig:construct_prompt2}
\end{figure*}


\end{document}